\documentclass[11pt]{article}
\usepackage{acl2015}
\usepackage{amsmath}
\usepackage{times}
\usepackage{url}
\usepackage{latexsym}
\usepackage{tikz}
\usetikzlibrary{fit,positioning}
\usepackage{caption}
\usepackage{subcaption}
\usepackage{amssymb}
\usepackage{amsmath}
\usepackage{algorithm}
\usepackage[noend]{algpseudocode}
\usepackage{footnote}
\usepackage{epstopdf}
\usepackage{setspace}
\usepackage{enumitem}
\usepackage{adjustbox}
\setlist[enumerate]{itemsep=0mm}
\usepackage[title]{appendix}
\usetikzlibrary{arrows}
\usepackage{refcount}
\usepackage{multirow}

\makeatletter
\newcommand{\@BIBLABEL}{\@emptybiblabel}
\newcommand{\@emptybiblabel}[1]{}
\makeatother
\usepackage{hyperref}
\allowdisplaybreaks

\newcommand{\note}[1]{}
\usepackage[disable]{todonotes} 

\newcommand*{\Scale}[2][4]{\scalebox{#1}{$#2$}}
\title{Fast, Flexible Models for Discovering Topic Correlation across Weakly-Related Collections}

\author{Jingwei Zhang$^1$, Aaron Gerow$^2$, Jaan Altosaar$^3$, James Evans$^{2,4}$, Richard Jean So$^5$ \\
  \\
  $^1$Department of Computer Science, Columbia University \\
  {\tt jz2541@columbia.edu} \\
  $^2$ Computation Institute, University of Chicago \\
  {\tt \{gerow,jevans\}@uchicago.edu} \\
  $^3$ Department of Physics, Princeton University \\
  {\tt altosaar@princeton.edu} \\
  $^4$ Department of Sociology, University of Chicago \\
  $^5$ Department of English Language and Literature, University of Chicago \\
  {\tt richardjeanso@uchicago.edu} \\
}

\date{}

\begin{document}
\maketitle

\begin{abstract}
    Weak topic correlation across document collections with different numbers of topics in individual collections presents challenges for existing cross-collection topic models.
    This paper introduces two probabilistic topic models, \textit{Correlated LDA} (C-LDA) and \textit{Correlated HDP} (C-HDP). These address problems that can arise when analyzing large, asymmetric, and potentially weakly-related collections. 
    Topic correlations in weakly-related collections typically lie in the tail of the topic distribution, where they would be overlooked by models unable to fit large numbers of topics.
    To efficiently model this long tail for large-scale analysis, our models implement a parallel sampling algorithm based on the Metropolis-Hastings and alias methods \cite{yuan2014lightlda}.
    The models are first evaluated on synthetic data, generated to simulate various collection-level asymmetries. We then present a case study of modeling over 300k documents in collections of sciences and humanities research from JSTOR.

\end{abstract}

\section{Introduction}
Comparing large text collections is a critical task for the
curation and analysis of human cultural history. Achievements of research
and scholarship are most accessible through textual artifacts, which
are increasingly available in digital archives. Text-based research,
often undertaken by humanists, historians, lexicographers, and
corpus linguists, explores patterns of words in documents
across time-periods and distinct collections of text. Here, we
introduce two new topic models designed to compare large collections, Correlated LDA (C-LDA) and Correlated HDP
(C-HDP), which are sensitive to document-topic asymmetry (where collections have different topic distributions)
and topic-word asymmetry (where a single topic has
different word distributions in each collection).
These models seek to address terminological questions, such as how
a topic on physics is articulated distinctively in scientific
compared to humanistic research.
Accommodating potential collection-level asymmetries is
particularly important when researchers seek to analyze collections with little
prior knowledge about shared or collection-specific topic structure.
Our models extend existing cross-collection
approaches to accommodate these asymmetries and implement an efficient
parallel sampling algorithm enabling
users to examine the long tail of topics in particularly large collections.

Using topic models for comparative text mining was introduced by
\newcite{zhai2004cross}, who developed the ccMix model which extended pLSA
\cite{hofmann1999probabilistic}. Later work by \newcite{paul2009cross}
developed ccLDA, which adopted the hierarchical Bayes framework
of Latent Dirichlet Allocation or LDA \cite{blei2003latent}. These models
account for topic-word asymmetry by assuming variation in the vocabularies of topics
is due to collection-level differences. Nevertheless, they require the same topics to be present in each
collection. These models are useful for comparing collections under specific
assumptions, but cannot accommodate
collection-topic asymmetry (which arises in collections that do not share every topic or
that have different numbers of topics). In situations where collections
do not share all topics, the results often include junk, mixed, or sparse topics,
making them difficult to interpret \cite{paul2009cross}.
Such asymmetries make it difficult to use models like ccLDA and ccMix when
little is known about collections in advance. This motivates our efforts to model variation in
the long tail of topic distributions, where correlations are more likely
to appear when collections are weakly related.

C-LDA and C-HDP extend ccLDA \cite{paul2009cross} to accommodate
collection-topic level asymmetries, particularly by allowing non-common topics
to appear in each collection. This added flexibility allows our
models to discover topic correlations across arbitrary collections with different
numbers of topics, even when there are few (or unknown) numbers of common topics.
To demonstrate the effectiveness of our models, we evaluate them
on synthetic data and show that they outperform related models such as ccLDA and differential topic models \cite{ding2014differential}.
We then fit C-LDA to two large collections of humanities and sciences
documents from JSTOR. Such historical analyses of text would be intractable without
an efficient sampler. An optimized sampler is required in such situations because common
topics in weakly-correlated collections are usually found in the tail of the
document-topic distribution of a sufficiently large set of topics.
To make this feasible on large datasets such as JSTOR, we employ a
parallelized Metropolis-Hastings \cite{kronmal} and alias-table sampling framework, adapted from LightLDA \cite{yuan2014lightlda}. These optimizations, which achieve
$\mathcal{O}(1)$ amortized sampling time per token, allow our models to be fit to
large corpora with up to thousands of topics in a matter of hours ---
an order of magnitude speed-up from ccLDA.

After reviewing work related to topic modeling across collections,
section \ref{model} describes C-LDA and C-HDP, and then details their technical
relationship to existing models. Section \ref{experiments} introduces
the synthetic data and part of the JSTOR corpus used in our evaluations.
We then compare our models' performances to other models in terms of held-out
perplexity and a measure of distinguishability.
The final results section exemplifies the use of C-LDA in a qualitative analysis of humanities and sciences research. We conclude with a brief discussion of the strengths of C-LDA and C-HDP, and outline directions for future work and applications.

\section{Related Work}
\label{related_work}
Our models seek to enable users to compare large collections that may only be weakly correlated and that may contain different numbers of topics. While topic models could be fit to separate collections to make post-hoc comparisons \cite{Wallach1,yang2011topic}, our goal is to account for both document-topic asymmetry and topic-word asymmetry ``in-model''.
In short, we seek to model the correlation between \textit{arbitrary} collections. 
Prioritizing in-model solutions for document-topic asymmetry has been explored elsewhere, such as in hierarchical Dirichlet processes (HDP), which use an additional level to account for collection variations in document-topic distributions \cite{teh2006hierarchical}.

One method designed to model topic-word asymmetry is ccMix \cite{zhai2004cross}, which models the generative probability of a word in topic $z$ from collection $c$ as a mixture of shared and collection-specific distributions $\theta_z$:
$$p(w) = \lambda_c\,p(w|\theta_z) + (1-\lambda_c) \,p(w|\theta_{z,c})$$
where $\theta_{z,c}$ is collection-specific and $\lambda_c$ controls the mixing between shared and collection-specific topics.
ccLDA extends ccMix to the LDA framework and adds a beta prior over $\lambda_c$ that reduces sensitivity to input parameters \cite{paul2009cross}.
%
Another approach, differential topic models~\cite{ding2014differential}, is based on hierarchical Bayesian models over topic-word distributions. This method uses the transformed Pitman-Yor process (TPYP) to model topic-word distributions in each collection, with shared common base measures.
As \cite{paul2009cross} note, ccLDA cannot accommodate a topic if it is not common across collections --- an assumption made by ccMix, ccLDA and the TPYP. In a situation where a topic is found in only one collection, it would either dominate the shared topic portion (resulting in a noisy, collection-specific portion), or it would appear as a mixed topic, revealing two sets of unrelated words \cite{Newman2010b}. C-LDA ameliorates this situation by allowing the number of common and non-common topics to be specified separately and by efficiently sampling the tail of the document-topic distribution, allowing users to examine less prominent regions of the topic space. C-HDP also grants collections document-topic independence using a hierarchical structure to model the differences between collections.

Due to increased demand for scalable topic model implementations, there has been a proliferation of optimized methods for efficient inference, such as SparseLDA \cite{yao2009efficient} and AliasLDA \cite{li2014reducing}.
%
%
AliasLDA achieves $\mathcal{O}(K_d)$ complexity by using the Metropolis-Hastings-Walker algorithm and an alias table to sample topic-word distributions in $\mathcal{O}(1)$ time. Although this strategy introduces temporal staleness in the updates of sufficient statistics, the lag is overcome by more iterations, and converges significantly faster.
A similar technique by \newcite{yuan2014lightlda}, LightLDA, employs cycle-based Metropolis Hastings mixing with alias tables for both document-topic and topic-word distributions. Despite introducing lag in the sufficient statistics, this method achieves $\mathcal{O}(1)$ amortized sampling complexity and results in even faster convergence than AliasLDA.
In addition to being fully parallelized, C-LDA adopts this sampling framework to make comparing large collections more tractable for large numbers of topics. Our models' efficient sampling methods allow users to fit large numbers of topics to big datasets where variation might not be observed in sub-sampled datasets or models with fewer topics.

\section{The Models}
\label{model}

\subsection{Correlated LDA}
In ccLDA (and ccMix), each topic has shared and collection-specific components for each collection.
C-LDA extends ccLDA to make it more robust with respect to topic asymmetries
between collections (Figure 1a).
The crucial extension is that by allowing each collection to define a set of
non-common topics in addition to common topics,
the model removes an assumption imposed by ccLDA and other
inter-collection models, namely that collections have the same number of topics.
As a result, C-LDA is suitable for collections without a large proportion
of common topics, and can also reduce noise (discussed in
Section~\ref{related_work}). To achieve this, C-LDA assumes
document $d$ in collection $c$ has a multinomial document-topic distribution
$\theta$ with an asymmetric Dirichlet prior for $K_c$ topics, where the
first $K^\emptyset$ are common across collections. It is also possible to introduce a tree structure
into the model that uses a binomial distribution to decide whether a word was drawn from
common or non-common topics. This yields collection-specific background topics by using
a binomial distribution instead of a multinomial. However, we prefer the simpler, non-tree version because
background topics are unnecessary when using an asymmetric $\alpha$ prior \cite{wallach2009rethinking}.


The generative process for C-LDA is as follows:

\begin{nobreak}
{\small
  \begin{enumerate}[leftmargin=4mm]
      \item Sample a distribution $\phi_k$ (shared component) from $\text{Dir}(\beta)$
      and a distribution $\sigma_{k}$ from $\text{Beta}(\delta_1, \delta_2)$
      for each common topic $k\in\{1,\ldots,K^\emptyset\}$;
  \item For each collection $c$,
      sample a distribution $\phi^c_k$ (collection-specific component) from $\text{Dir}(\beta)$ for
      each common topic $k\in\{1,\ldots,K^\emptyset\}$ and non-common
      topic $k\in\{K^\emptyset+1,\ldots,K_c\}$;
  \item For each document $d$ in $c$,
    sample a distribution $\theta$ from $\text{Dir}(\alpha_c)$;
  \item For each word $w_i$ in $d$:
    \begin{enumerate}[leftmargin=6mm]
    \item Sample a topic $z_i\in\{1,\ldots,K_c\}$ from $\text{Multi}(\theta)$;
    \item If $z_i\leq K^\emptyset$, sample $y_i$ from $\text{Binomial}(\sigma_{z_i})$;
    \item Sample $w_i$ from $\text{Multi}(\phi^\xi_{z_i})$, where\\
      $\xi=\left\{\begin{array}{ll}\mathtt{null}&,z_i\leq K^\emptyset  \text{ and } y_i=0;\\c&\text{, otherwise.}\end{array}\right.$
    \end{enumerate}
  \end{enumerate}
}
\end{nobreak}
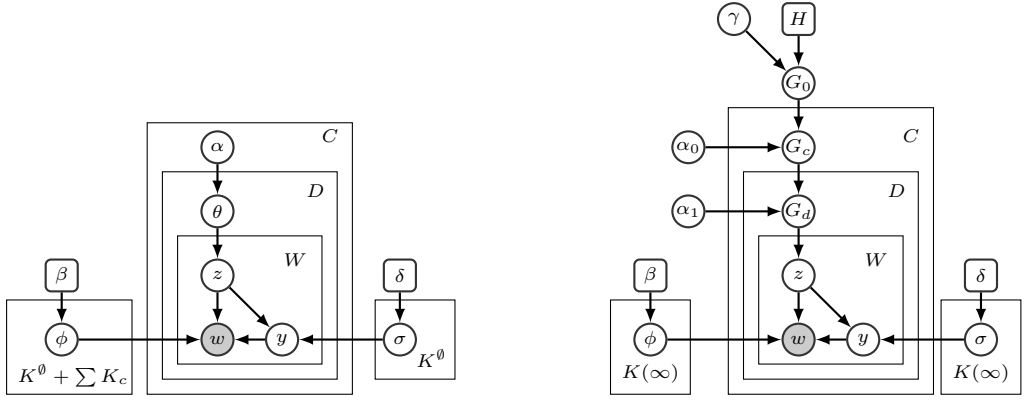
\begin{figure*}[t]
\centering
\begin{subfigure}[b]{0.5\textwidth}
  \begin{center}
    \scriptsize
    \centerline{\begin{tikzpicture}
        \tikzstyle{main}=[circle, inner sep=0pt, minimum size=12pt, thick, draw =black!80, node distance = 4mm]
        \tikzstyle{param}=[inner sep=0pt, minimum size=12pt, thick, draw =black!80, node distance = 4mm, rounded corners=2pt]
        \tikzstyle{connect}=[-latex, thick]
        \tikzstyle{box}=[rectangle, draw=black!100]
        \node[main] (alpha) [] {$\alpha$};
        \node[main] (theta) [below=of alpha] {$\theta$};
        \node[main] (z) [below=of theta] {$z$};
        \node[main, fill=black!20] (w) [below=of z] {$w$};
        \node[main] (y) [right=of w] {$y$};
        \node[main] (phi) [left=1.6cm of w] {$\phi$};
        \node[main] (sigma) [right=1.1cm of y] {$\sigma$};
        \node[param] (delta) [above=of sigma] {$\delta$};
        \node[param] (beta) [above=of phi] {$\beta$};
        \path (delta) edge [connect] (sigma)
        (beta) edge [connect] (phi)
        (alpha) edge [connect] (theta)
        (theta) edge [connect] (z)
        (sigma) edge [connect] (y)
        (z) edge [connect] (y)
        (z) edge [connect] (w)
        (y) edge [connect] (w)
        (phi) edge [connect] (w);
        \node[rectangle, inner xsep=3mm, inner ysep=2mm, yshift=1mm, draw=black!100, fit= (z) (y)] {};
        \node[rectangle, inner sep=0mm, fit= (z) (y),label=above right:{$W$}, yshift=-2mm, xshift=-3mm] {};
        \node[rectangle, inner ysep=3mm, inner xsep=5mm, xshift=0mm,draw=black!100, fit= (theta) (y)] {};
        \node[rectangle, inner ysep=3mm, yshift=-2mm, inner xsep=7mm, xshift=0mm, draw=black!100, fit= (alpha) (y)] {};
        \node[rectangle, inner sep=0mm, fit= (theta) (y),label=above right:{$D$}, yshift=3mm, xshift=0mm] {};
        \node[rectangle, inner sep=0mm, fit= (alpha) (y),label=above right:{$C$}, yshift=6mm, xshift=2mm] {};
        \node[rectangle, inner sep=3mm, xshift=2mm, draw=black!100, fit= (sigma)] {};
        \node[rectangle, inner sep=0mm, fit= (sigma),label=below right:{$K^\emptyset$}, yshift=2mm, xshift=-1mm] {};
        \node[rectangle, inner ysep=4mm, inner xsep=6mm, yshift=-1mm, xshift=1mm, draw=black!100, fit= (phi)] {};
        \node[rectangle, inner sep=0mm, fit= (phi) ,label=below left:{$K^\emptyset+\sum K_c$}, yshift=0mm, xshift=12mm] {};
    \end{tikzpicture}}
    \label{fig:bptf}
  \end{center}
\end{subfigure}
\begin{subfigure}[b]{0.45\textwidth}
  \begin{center}
    \scriptsize
    \centerline{\begin{tikzpicture}
        \tikzstyle{main}=[circle, inner sep=0pt, minimum size=12pt, thick, draw =black!80, node distance = 4mm]
        \tikzstyle{param}=[inner sep=0pt, minimum size=12pt, thick, draw =black!80, node distance = 4mm, rounded corners=2pt]
        \tikzstyle{connect}=[-latex, thick]
        \tikzstyle{box}=[rectangle, draw=black!100]
        \node[main] (gc) [] {$G_c$};
        \node[main] (g0) [above=of gc] {$G_0$};
        \node[param] (h) [above=of g0] {$H$};
        \node[main] (gamma) [left=of h] {$\gamma$};
        \node[main] (theta) [below=of gc] {$G_d$};
        \node[main] (alpha0) [left=1cm of gc] {$\alpha_0$};
        \node[main] (alpha1) [left=1cm of theta] {$\alpha_1$};
        \node[main] (z) [below=of theta] {$z$};
        \node[main, fill=black!20] (w) [below=of z] {$w$};
        \node[main] (y) [right=of w] {$y$};
        \node[main] (phi) [left=1.5cm of w] {$\phi$};
        \node[main] (sigma) [right=1.1cm of y] {$\sigma$};
        \node[param] (delta) [above=of sigma] {$\delta$};
        \node[param] (beta) [above=of phi] {$\beta$};
        \path (delta) edge [connect] (sigma)
        (beta) edge [connect] (phi)
        (gc) edge [connect] (theta)
        (g0) edge [connect] (gc)
        (h) edge [connect] (g0)
        (gamma) edge [connect] (g0)
        (alpha0) edge [connect] (gc)
        (alpha1) edge [connect] (theta)
        (theta) edge [connect] (z)
        (sigma) edge [connect] (y)
        (z) edge [connect] (y)
        (z) edge [connect] (w)
        (y) edge [connect] (w)
        (phi) edge [connect] (w);
        \node[rectangle, inner xsep=3mm, inner ysep=2mm, yshift=1mm, draw=black!100, fit= (z) (y)] {};
        \node[rectangle, inner sep=0mm, fit= (z) (y),label=above right:{$W$}, yshift=-2mm, xshift=-3mm] {};
        \node[rectangle, inner ysep=3mm, inner xsep=5mm, xshift=0mm,draw=black!100, fit= (theta) (y)] {};
        \node[rectangle, inner ysep=4mm, yshift=-1mm, inner xsep=7mm, xshift=0mm, draw=black!100, fit= (gc) (y)] {};
        \node[rectangle, inner sep=0mm, fit= (theta) (y),label=above right:{$D$}, yshift=3mm, xshift=0mm] {};
        \node[rectangle, inner sep=0mm, fit= (gc) (y),label=above right:{$C$}, yshift=6mm, xshift=2mm] {};
        \node[rectangle, inner ysep=4mm, inner xsep=3mm, yshift=-1mm, draw=black!100, fit= (sigma)] {};
        \node[rectangle, inner sep=0mm, fit= (sigma),label=below:{$K(\infty)$}, yshift=0mm, xshift=0mm] {};
        \node[rectangle, inner ysep=4mm, inner xsep=3mm, yshift=-1mm, xshift=0mm, draw=black!100, fit= (phi)] {};
        \node[rectangle, inner sep=0mm, fit= (phi) ,label=below:{$K(\infty)$}, yshift=0mm, xshift=0mm] {};
    \end{tikzpicture}}
    \label{fig:bptf}
  \end{center}
\end{subfigure}
\caption{Graphical models of C-LDA (a; left) and C-HDP (b; right).}
\label{fig:model}
\end{figure*}
Note that to capture common
topics, $K^\emptyset$ should be set such that $\exists~c$ where $K_c=K^\emptyset$.
Otherwise, words sampled as a non-common topic will not have
information about non-common topics in other collections. Then a
``common-topic word'' is found among non-common topics in all collections (a
local minima) and it will take a long time to stabilize as a common topic.
To avoid this, when determining the number of topics for sampling, the
number of non-common topics for the collection with the smallest number of total
topics should be zero.
After inference, to distinguish common and non-common topics in this
collection, we model $\sigma$ independently by assuming
collections have the same mixing ratio for common topics. With this reasonable
assumption and an asymmetric $\alpha$, common topics become sparse enough that
some $\sigma$ distributions reduce nearly to 0, distinguishing them as non-common
topics. Although this may seem counterintuitive, it does not negatively affect
results.


Three kinds of collection-level imbalance can confound inter-collection topic models:
1) in the numbers of topics between collections,
2) in the numbers of documents between collections, and
3) in the document-topic distributions.
Each of these can cause topics in different collections to have
significantly different numbers of words assigned to the same topic. In this way,
a topic can be dominated by the collection comprising most of its words.
C-LDA addresses imbalances in the document-topic distributions between collections
by estimating $\alpha$.
For imbalance in the number of topics and documents, C-LDA mimics document over-sampling
in the Gibbs sampler using a different
unit-value in the word count table for each collection.
Specifically, a unit $\eta_c$ is chosen for each collection such that the
average equivalent number of assigned words per-topic ($\sum_{d\in c}
\eta_c N_d/K_c$, where $N_d$ is the length of document $d$) is equal.
This process both increases the topic quality (in terms of
collection balance) in the resulting held-out perplexity of the model.



\subsection{Correlated HDP}
To alleviate C-LDA's requirement that $\exists~c$ such that
$K_c=K^\emptyset$, we introduce a variant of the model, the correlated hierarchical Dirichlet process (C-HDP), that uses a 3-level hierarchical Dirichlet process \cite{teh2006hierarchical}. The generative process for C-HDP is the same
as C-LDA shown above, except that here we assume a word's topic, $z$, is generated by a hierarchical Dirichlet
process:
$$
\begin{array}{rcl}
G_0|\gamma,H&\sim&\text{DP}(\gamma,H)\\
G_c|\alpha_0,G_0&\sim&\text{DP}(\alpha_0,G_0)\\
G_d|\alpha_1,G_c&\sim&\text{DP}(\alpha_1,G_c)\\
z|G_d&\sim&G_d
\end{array}
$$

\noindent
where $G_0$ is a base measure for each collection-level Dirichlet process, and $G_c$ are base measures of
document-level Dirichlet processes in each collection (Figure 1b). Thus, documents from the same
collection will have similar topic distributions compared to those from
other collections, and collections are allowed to have distinct sets of topics due to the use of HDP.


\section{Inference}
\label{inference}

\subsection{Posterior Inference in C-LDA}
\label{sec:inference}
C-LDA can be trained using collapsed Gibbs sampling with $\phi$,
$\theta$, and $\sigma$ integrated out. Given the status assignments of other words,
the sampling distribution for word $w_i$ is given by:
\begin{equation}
  \label{eq1}
  \begin{split}
    ~&p(y_i,z_i|\boldsymbol w,\boldsymbol y_{-i},\boldsymbol z_{-i},\delta,\alpha,\beta)\\
    \propto~&\underbrace{(N(d,z_i)+\alpha_{c,z_i})}_{q_d}\\
    ~&\times\underbrace{\left\{\setstretch{2}\Scale[0.8]{\begin{array}{ll}
          \dfrac{N(y_i,z_i)+\delta_{y_i}}{N(z_i)+\sum_k\delta_{k}}\times\dfrac{N(w_i, y_i, z_i, \zeta)+\beta}{N(y_i, z_i, \zeta)+V\beta}&z_i\leq K^\emptyset\\
          \dfrac{N(w_i,
            z_i,c)+\beta}{N(z_i,c)+V\beta}&z_i>K^\emptyset\end{array}}\right.}_{q_w}
  \end{split}
\end{equation}
\noindent
where $\zeta=$ {\small $\left\{\begin{array}{ll}*&y_i=0\\c&y_i=1\end{array}\right.$ },
$N(\cdots)$ is the number of status assignments for
$(\cdots)$, not including $w_i$.

Inference in C-LDA employs two optimizations: a parallelized
sampler and an efficient sampling algorithm (Algorithm~\ref{alg}).
We use the parallel schema in \cite{smola2010architecture,lu2013accelerating} which applies atomic updates to the sufficient statistics to avoid race conditions.
The key idea behind the optimized sampler is the combination of alias tables and the
Metropolis-Hastings method (MH), adapted from \cite{yuan2014lightlda,li2014reducing}.
Metropolis-Hastings is a Markov chain Monte Carlo method that uses a proposal distribution to approximate
the true distribution when exact sampling is difficult.
In a complimentary way, Walker's alias method \shortcite{marsaglia2004fast} allows one to
effectively sample from a discrete distribution by using an alias table,
constructed in $\mathcal{O}(K)$ time, from which we can sample in
$\mathcal{O}(1)$. Thus, reusing the sampler $K$ times as the proposal
distribution for Metropolis-Hastings yields $\mathcal{O}(1)$ amortized sampling time per-token.

\begin{algorithm}[t]
  \caption{Sampling in C-LDA}
  \label{alg}
  {\small
    \begin{algorithmic}
      \Repeat
      \For{all documents $\{d\}$ in parallel}
      \For{words $\{w\}$ in $d$}
      \State $z\leftarrow$ \Call{CycleMH}{$p,q_w,q_d,z$}
      \State sample $y$ given $z$
      \EndFor
      \State Atomic update sufficient statics
      \EndFor
      \State Estimate $\alpha$
      \Until convergence
      \Statex
      \Procedure{CycleMH}{$p,q_w,q_d,z$}
      \For{$i=1$ {\bfseries to} $N$}
      \If {$i$ is even}
      \State proposal $q\leftarrow q_w$
      \Else
      \State proposal $q\leftarrow q_d$
      \EndIf
      \State sample $z'\sim$ \Call{AliasTable}{$q$}
      \If {$\text{RandUnif(1)}<\min(1,\frac{p(z')q(z)}{p(z)q(z')})$}
      \State $z\leftarrow z'$
      \EndIf
      \EndFor
      \Return $z$
      \EndProcedure
    \end{algorithmic}
  }
\end{algorithm}

Notice that in Eq.~\ref{eq1}, the sampling distribution is the product of
a single document-dependent term $q_d$ and a single word-dependent term $q_w$. After burn-in,
both terms will be sparse (without the smoothing factor). It is
therefore reasonable to use $q_d$ and $q_w$ as cycle proposals
\cite{yuan2014lightlda}, alternating them in each Metropolis-Hastings step.
Our experiments show that the primary drawback of this method --- stale sufficient
statistics --- does not empirically affect convergence.
Our implementation uses proposal distributions $q_w$ and $q_d$, with $y$
marginalized out. After the Metropolis-Hastings steps, $y$ is sampled to update $z$, to
reduce the size of the alias tables, yielding even faster convergence.

Lastly, the use of an asymmetric $\alpha$ allows C-LDA to discover
correlations between less dominant topics across collections \cite{wallach2009rethinking}.
We use Minka's fixed-point method, with a gamma hyper-prior
to optimize $\alpha_c$ for each collection separately \cite{wallach2008structured}.
All other hyperparameters were fixed during inference.

\subsection{Posterior Inference in C-HDP}
C-HDP uses the block sampling algorithm described in \cite{chen2011sampling},
which is based on the Chinese restaurant process metaphor.
Here, rather than tracking all assignments (as the samplers given in \cite{teh2006hierarchical}),
table indicators are used to track only the start of new tables,
which allows us to adopt the same sampling framework as C-LDA.
In the Chinese restaurant process, each Dirichlet process in the hierarchical structure
is represented as a restaurant with an infinite number of tables, each serving the same dish.
New customers can either join a table with existing customers, or start a new table.
If a new table is chosen, a proxy customer will be sent to
the parent restaurant to determine the dish served to that table.

In the block sampler, indicators are used to denote a customer creating a table (or tables) up to level $u$
(0 as the root, 1 for collection level, and 2 for the document level), and
$u=\emptyset$ indicates no table has been created.
For example, when a customer creates a table at the collection level, and the proxy customer in the collection level creates a table at the root level, $u$ is $0$.
With this metaphor, let $n_{lz}$ be the number of customers (including their proxies) served dish $z$ at restaurant $l$, and let $t_{lz}$ be the number of tables serving dish $z$ at restaurant $l$ ($l=0$ for root, $l=c$ for collection level or $l=d$ for document level), with $N_0=\sum_z n_{0z}$ and $N_c=\sum_z n_{cz}$. By the chain rule, the conditional probability of the state assignments for $w_i$, given all others, is

\vspace{-0.5cm}
{\small
\begin{align*}
&p(y_i,z_i,u_i|\boldsymbol w,\boldsymbol y_{-i},\boldsymbol z_{-i},\boldsymbol u_{-i},\ldots)\\
\propto&\frac{N(y,z)+\delta_{y}}{N(z)+\sum_k\delta_{k}}\times\frac{N(w, y, z,\zeta)+\beta}{N(y, z, \zeta)+V\beta}\\
\times&
\left\{\begin{array}{ll}
    \frac{\gamma\alpha_0}{\gamma+N_0} & u=0\\
    \frac{\alpha_0}{\gamma+N_0}\frac{S^{n_{cz}+1}_{t_{cz}+1}}{S^{n_{cz}}_{t_{cz}}}\frac{S^{n_{dz}+1}_{t_{dz}+1}}{S^{n_{dz}}_{t_{dz}}}\frac{n_{0z}^2(t_{cz}+1)(t_{dz}+1)}{(n_{0z}+1)(n_{cz}+1)(n_{dz}+1)} & u=1\\
    \frac{S^{n_{cz}+1}_{t_{cz}}}{S^{n_{cz}}_{t_{cz}}}\frac{S^{n_{dz}+1}_{t_{dz}+1}}{S^{n_{dz}}_{t_{dz}}}\frac{(t_{dz}+1)(n_{cz}-t_{cz}+1)}{(n_{cz}+1)(n_{dz}+1)} & u=2\\
    \frac{\alpha_0+N_1}{\alpha_1}\frac{S^{n_{dz}+1}_{t_{dz}}}{S^{n_{dz}}_{t_{dz}}}\frac{n_{dz}-t_{dz}+1}{n_{dz}+1} & u=\emptyset
\end{array}\right.
\end{align*}
}

\noindent
Here, $S_t^n$ is the Stirling number, the ratios of which can be efficiently precomputed
\cite{buntine2010bayesian}. The concentration parameters $\gamma, \alpha_0$, and
$\alpha_1$ can be sampled using the auxiliary variable method \cite{teh2006hierarchical}.

\begin{figure*}[t]
  \begin{subfigure}[t]{0.49\textwidth}
    \centering
    \includegraphics[width=.75\linewidth]{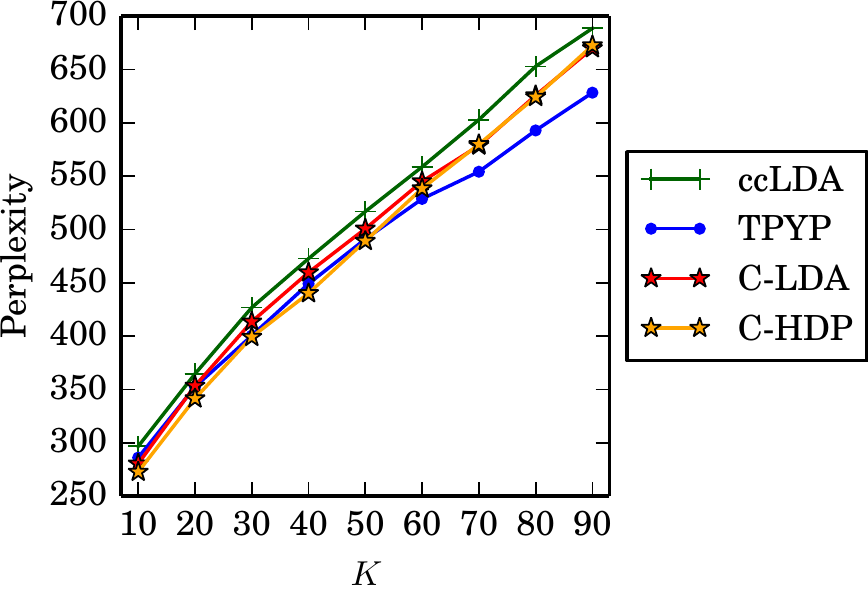}
  \end{subfigure}
  \begin{subfigure}[t]{0.49\textwidth}
    \centering
    \includegraphics[width=.85\linewidth]{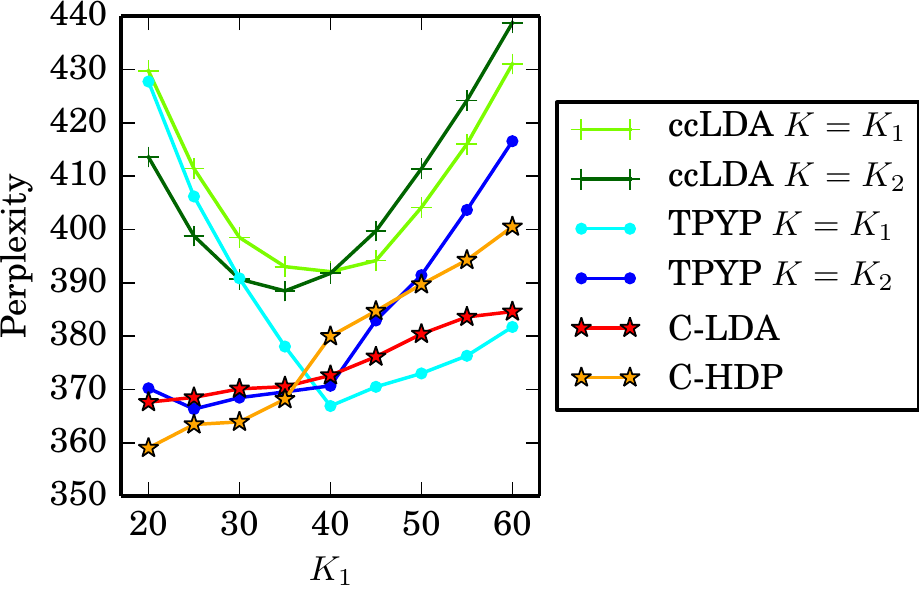}
  \end{subfigure}
  \caption{Held-out perplexity of C-LDA, C-HDP, ccLDA and TPYP fit to synthetic data, where $K_1=K_2=K$ (a; left) and data with an asymmetric number of topics (b; right).}
  \label{fig:exp}
\end{figure*}

Note that because conditional probability has the same separability as
C-LDA (to give term $q_w$ and $q_d$), the same sampling framework can be used with two alterations: 1) when a new topic is created or removed at the
root, collection, or document level, the related alias tables must be reset, which makes the sampling slightly slower than $\mathcal{O}(1)$, and 2) while the document alias table samples $z$ and $u$ simultaneously, after sampling $z$ from the word alias table $u$ must be sampled using $t_{lc}/n_{lz}$~\cite{chen2011sampling}.
Parallelizing C-HDP requires an additional empirical method of merging new topics between threads \cite{newman2009distributed}, which is outside of the scope of this work.
Our implementation of both models, C-LDA and C-HDP, are open-sourced online
\footnote{\href{https://github.com/iceboal/correlated-lda}{https://github.com/iceboal/correlated-lda}}.

\section{Experiments}
\label{experiments}

\subsection{Model Comparison}
We use perplexity on held-out documents to evaluate the performance of C-LDA and C-HDP.
In all experiments, the gamma prior for $\alpha$ in C-LDA was set to $(1,1)$, and $(5,0.1)$, $(5,0.1)$, $(0.1,0.1)$ for $\gamma$, $\alpha_0$, $\alpha_1$ respectively in C-HDP.
In the hold-out procedure, 20\% of documents were randomly selected as test data.
LDA, C-LDA and ccLDA were run for 1,000 iterations and C-HDP and the TII-variant of TPYP for 1,500 iterations (unless otherwise noted), all of which converged to a state where change in perplexity was less than 1\% for ten consecutive iterations.

Perplexity was calculated from the marginal likelihood of a held-out document
$p(\mathbf{w}|\mathbf{\Phi},\alpha)$, estimated using the ``left-to-right''
method \cite{Wallach09}. Because it is difficult to validate real-world data that exhibits different kinds of asymmetry, we use
synthetic data generated specifically
for our evaluation tasks \cite{AlSumait,Wallach09,kucukelbir2014profile}.

\subsubsection{Topic Correlation}
C-LDA is unique in the amount of freedom it allows when setting the number of topics for collections.
To assess the models' performances with various topic correlations in a fair setting,
we generated two collections of
 synthetic data by following the generative process (varying
the number of topics) and measured the models' perplexities against the ground truth parameters.
In each experiment, two collections were generated, each with 1,000 documents containing 50 words each, over a vocabulary of 3,000. $\beta$ and $\delta$ were fixed at 0.01 and 1.0 respectively, and $\alpha$
was asymmetrically defined as $1/(i+\sqrt{K_c})$ for $i\in[0,K_c-1]$.


\paragraph{Completely shared topics}
The assumptions imposed by ccLDA and TPYP effectively make them a special case of our
model where $K^\emptyset=K_1=K_2=\ldots$. To compare results, data was generated
such that all numbers of topics were equal to $K\in[10,90]$.
Additionally, all models were configured to use this ground truth parameter when training.
Not surprisingly, ccLDA, C-LDA, and C-HDP have almost the same perplexity with respect to $K$
because their structure is the same when all topics are shared (Figure~\ref{fig:exp}a).

\paragraph{Asymmetric numbers of topics}
To explore the effect of asymmetry in the number of topics, data was generated such that one collection had
$K_1\in[20,60]$ topics while a second had a fixed $K_2=40$ topics.
The number of shared topics was set to $K^\emptyset=20$.
The parameters for C-LDA and C-HDP (initial values) were set to ground truths, and, to retain a fair comparison, versions of ccLDA and TPYP were fit with both $K=K_1$ and $K=K_2$.

We find that ccLDA performs nearly as well as C-LDA and C-HDP when there is more symmetry between collection, namely when $K_1\approx K_2$ (Figure~\ref{fig:exp}b).
TPYP, on the other hand, performs well with more topics ($2\times\max(K_1,K_2)$ where the ground truth is $K_1~\&~K_2$).
In contrast, C-LDA and C-HDP perform more consistently than other models across varying degrees of asymmetry.

\paragraph{Partially-shared topics}
When collections have the same number of topics,
C-LDA, C-HDP and ccLDA exhibit adequate flexibility, resulting in similar perplexities.
When collections have increasingly few common topics, however,
common and non-common topics from ccLDA
are considerably less distinguishable than those from C-LDA. To evaluate the models'
abilities in such situations, data was generated for two
collections having $K_1=K_2=50$ topics, but with the shared number of topics $K^\emptyset \in [5,45]$.
We also set $\delta^{(0)}=\delta^{(1)}=5$, and for comparison to ccLDA we used $K=50$.

\begin{figure}[h]
    \centering
    \includegraphics[width=.7\linewidth]{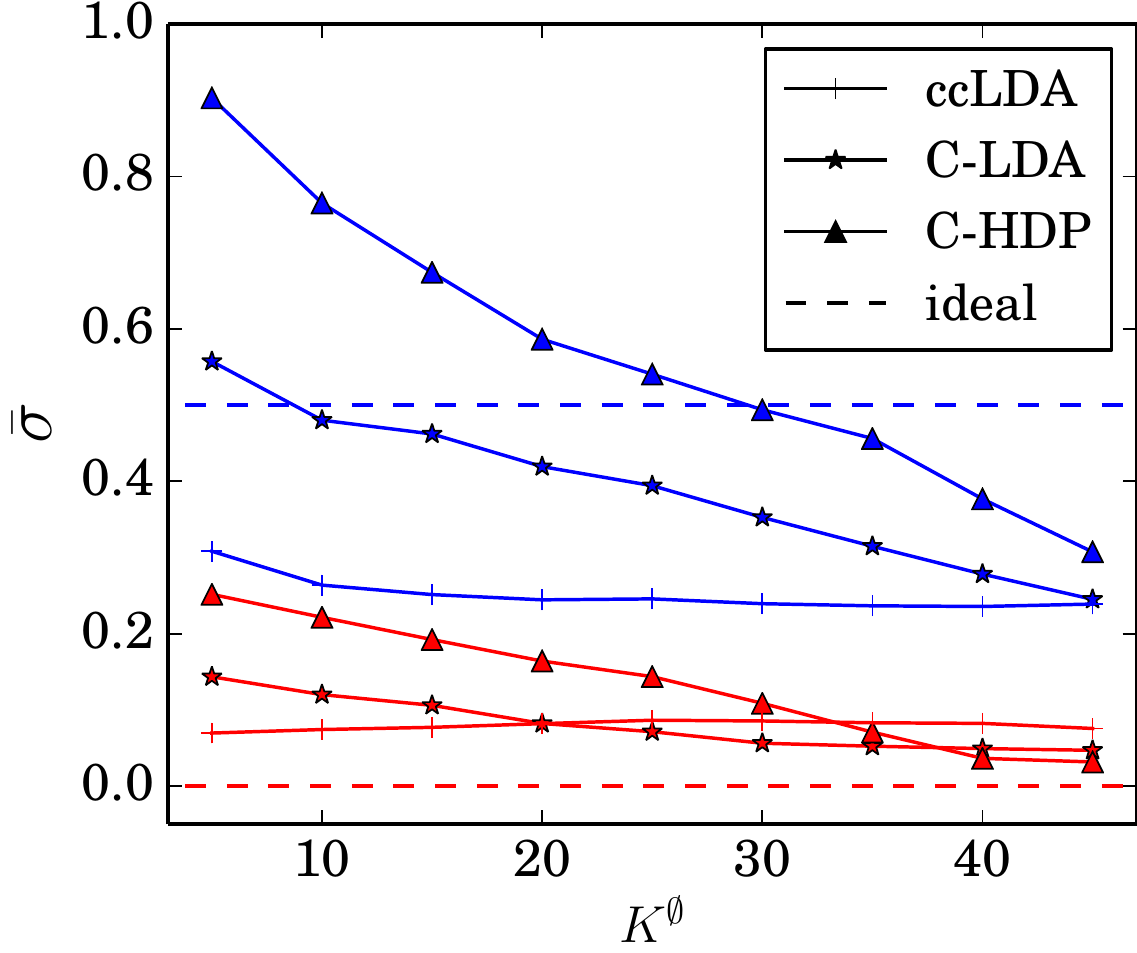}
    \caption{Distinguishability (Eq. \ref{eq:interp}) of topics fit with C-LDA,C-HDP and ccLDA. Blues lines denote $\bar{\sigma}_{\text{common}}$ and red denote $\bar{\sigma}_{\text{non-common}}$.}
    \label{fig:exp2}
\end{figure}

To measure this distinguishability, we examine the inferred $\sigma$.
Recall that $\sigma$ indicates what percentage of a common topic is shared. When a topic is actually non-common, the value of $\sigma$ should be small.
We sort $\sigma_k$ for $k\in[1,K]$ in reverse and use
\begin{equation}
\begin{array}{r l}
    \bar{\sigma}_{\text{common}}=&\frac{1}{K^\emptyset}\sum_{k=1}^{K^\emptyset}\sigma_k\\
    \\
    \bar{\sigma}_{\text{non-common}}=&\frac{1}{K-K^\emptyset}\sum_{k=K^\emptyset+1}^{K}\sigma_k
\end{array}
\label{eq:interp}
\end{equation}
as measures of how well common and non-common topics were learned\footnote{TPYP is not comparable using this metric, but its hierarchical structure will cause topics to mix naturally.}.
$\bar{\sigma}_{\text{common}}$ is the average of the $K^\emptyset$ largest $\sigma$ values, and
$\bar{\sigma}_{\text{non-common}}$ is the average of the rest.
When $\delta^{(0)}=\delta^{(1)}$ in the synthetic data, $\sigma$ in the common portion
should be 0.5, whereas it should be 0 in the non-common part.
Figure~\ref{fig:exp2} shows that C-LDA better distinguishes between common and
non-common topics, especially when $K^\emptyset$ is small. This allows non-common topics to be separated from
the results by examining the value of $\sigma$. C-HDP has similar performance but larger $\sigma$ values. In ccLDA, all topics are shared between
collections which means that common and non-common topics are mixed.
As expected, ccLDA performs similarly when all topics are common across collections.

\begin{figure*}[th]
  \begin{tabular}{lr}
    \centering
    \includegraphics[scale=.45]{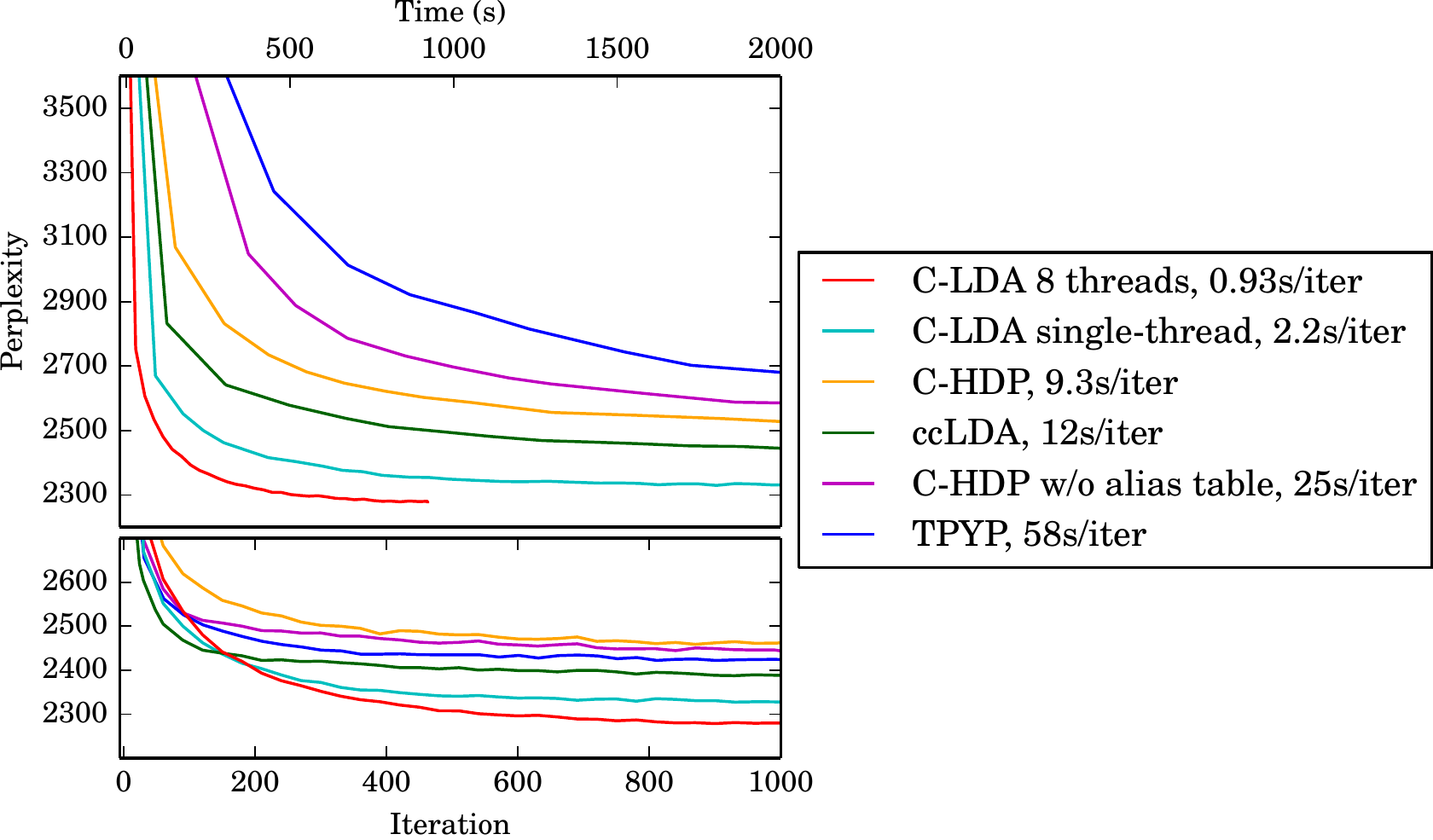}
    &
    \includegraphics[scale=.75]{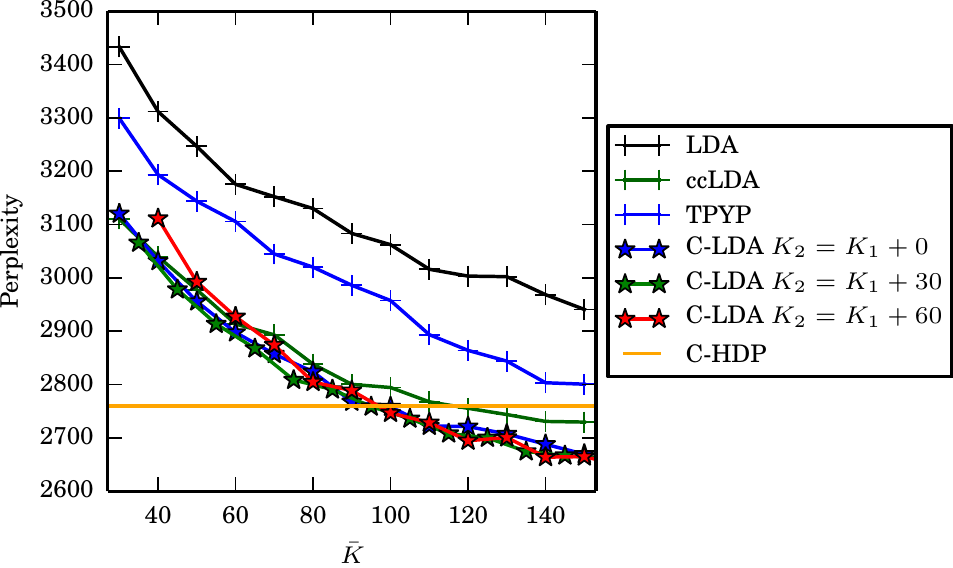}
  \end{tabular}
  \vspace{-3mm}
  \caption{Using JSTOR: perplexity vs. runtime and iterations (a; left) and perplexity vs. $K$ (b; right).}
  \label{fig:perp}
\end{figure*}

\begin{table*}[t]
  \centering
  \small
  \begin{tabular}{|c|rrr|rr|rr|}
    \hline
    &\multicolumn{5}{|c}{\textbf{Coherence}} & \multicolumn{2}{|c|}{\textbf{Mutual Coherence}}\\
    &\multicolumn{3}{|c}{shared component} & \multicolumn{2}{c|}{collection-specific} & \multicolumn{2}{c|}{shared \& collection-specific}\\
    \hline
    &all documents & science & humanities & science & humanities & science & humanities \\
    \hline
    C-LDA & -8.83 & -7.73 & -8.04 & -8.38 & -8.14 & -8.54 & -8.37\\
    ccLDA & -9.04 & -8.22 & -8.27 & -8.38 & -8.15 & -8.69 & -8.40\\
    \hline
    C-LDA & -7.22 & -3.68 & -6.11 & -8.25 & -8.09 & -7.75 & -7.97\\
    ccLDA & -8.11 & -5.68 & -7.12 & -8.24 & -7.88 & -8.22 & -7.95\\
    \hline
  \end{tabular}
  \vspace{-2mm}
  \caption{Average semantic coherence of the 50 common topics from JSTOR (top) and the average of the 10 best common topics judged by the mean value of different types of coherence (bottom).}
  \label{tab:coherence}
\end{table*}

\subsection{Semantic Coherence}
Semantic coherence is a corpus-based metric of the quality of a topic,
defined as the average pairwise similarity of the top $n$ words \cite{Newman2010,Mimno2011}.
A PMI-based form of coherence, which has been found to be the best proxy human judgements of topic quality,
 is defined for a topic $k$ as:
\begin{equation*}
\text{C}(k) = {{2} \over {n(n-1)}} {\sum \limits^n_{\substack{(w_i,w_j) \in k \\ i < j}}} \log {{D(w_i,w_j)+1} \over {D(w_i)D(w_j)}}
\label{eq:coherence}
\end{equation*}
\noindent
where $D(\cdot)$ computes the document co-occurrence. To accommodate coherence
with common topics in C-LDA that have shared and
collection-specific components we define \textit{mutual coherence}, $\text{MC}(k)$, as
\begin{equation*}
\text{MC}(k) = {{1} \over {n^2}} {\sum \limits^n_{\substack{w_i\in\text{shared},\\w_j\in\text{collection-specific}}}} \log {{D(w_i,w_j)+1} \over {D(w_i)D(w_j)}}
\label{eq:mcoherence}
\end{equation*}
so that for each collection, $\text{C}(k)$ ($2n$ words) is equal to
$\text{C}(k,\text{shared})$ + $\text{C}(k,\text{collection-specific})$ +
$\text{MC}(k)$.
Table~\ref{tab:coherence} shows the semantic coherence of topics fit with ccLDA and C-LDA. We used a 10\% sample of JSTOR due to the limited speed of ccLDA,
using 50 (common) topics for ccLDA / C-LDA, and 250 non-common humanities topics for C-LDA.
Although these settings are different for the models, the science topics are still comparable because they both have 50 topics. We found that C-LDA provides improved coherence in nearly all situations.

\subsubsection{Inference Efficiency}
To compare the model efficiency, we timed runs on a sample of 5,036 documents from JSTOR (introduced in the next section) with a 20\% held-out and set $K=K_1=K_2=200$ run on a commodity computer with four cores and 16GB of memory.
Figure~\ref{fig:perp}a shows the perplexity over time and iterations.
The inference algorithm introduces some staleness, which yields slower convergence in the first 200 iterations.
This effect, however, is outweighed in both C-LDA and C-HDP by the increased sampling speed.
With 8 threads, C-LDA not only converges faster, but yields lower perplexity, likely due to threads introducing additional stochasticity.



\subsection{Performance on JSTOR}
To compare our models against slower models, we sampled 2,465 documents from JSTOR, withholding 20\% as testing set.
We fit a model with 100 common and 50 non-common initial topics using C-HDP, which produced 272 root topics after 2,000 iterations.
The perplexity scores are roughly the same when C-LDA uses the same average number of topics per collection (Figure~\ref{fig:perp}b), except when numbers of topics are very asymmetric. Our model begins to outperform ccLDA after 80 topics. C-HDP did not, however, out-perform C-LDA despite the original HDP outperforming LDA. This could be do to the fact that the hierarchical structure of C-HDP is considerably different than the typical 2-level HDP.
Held-out perplexity on real data provides a quantitative evaluation of our models' performance in a real-world setting. However, the goal of our models is to enable a deeper analysis of large, weakly-related corpora, which we next discuss.

\begin{table*}[tmbh]
  \centering
  \small
  \begin{tabular}{|c|c|c||c|c|c||c|c|c|}
    \hline
    \multicolumn{3}{|c||}{{\bf Topic 2}} & \multicolumn{3}{c||}{{\bf Topic 21}} & \multicolumn{3}{c|}{{\bf Topic 23}}\\
    \hline
    {\bf shared} & {\bf science} & {\bf humanities} & {\bf shared} & {\bf science} & {\bf humanities} & {\bf shared} & {\bf science} & {\bf humanities} \\
    \hline
    animal & pig     & beast    & economic    & cost       & rural      & particle    & energy   & universe   \\
    specie & fly     & creature & government  & industry   & local      & physic      & electron & quantum    \\
    dog    & monkey  & nonhuman & economy     & company    & community  & physicist   & ray      & physic     \\
    wild   & guinea  & natural  & trade       & price      & village    & energy      & ion      & technical  \\
    wolf   & primate & humanity & major       & market     & region     & experiment  & atom     & scientific \\
    monkey & worm    & bird     & growth      & product    & urban      & event       & particle & relativity \\
    horse  & dog     & living   & capital     & income     & country    & measurement & mass     & physical   \\
    sheep  & cat     & gorilla  & industry    & industrial & area       & atom        & neutron  & mechanic   \\
    lion   & mammal  & brute    & institution & business   & regional   & interaction & proton   & law        \\
    cat    & cattle  & ape      & support     & private    & population & atomic      & nucleus  & reality    \\
    \hline
  \end{tabular}
  \vspace{-2mm}
  \caption{Three topics from the JSTOR collections with their top words in shared and specific components. \small{Complete results available at \href{http://j.mp/jstor-html}{http://j.mp/jstor-html}.}}
  \label{tab:topwords}
\end{table*}

\subsection{Qualitative Analysis}
\label{jstor}
Our models are designed to enable researchers to compare collections of text
in a way that is scalable and sensitive to collection-level asymmetries. To demonstrate that C-LDA can fill this
role, we fit a model to the entire JSTOR sciences and humanities collections
with 100 science topics and 1000 humanities topics (to reveal the less popular science-related topics in the humanities), and $\beta=0.01, \delta=1.0$.
JSTOR includes books and journal publications in over 9 million documents across
nearly 3 thousand journals.
We used the journal \textit{Science} to represent a collection of scientific research and 76 humanist
journals to represent humanities research\footnote{The list is available at http://j.mp/humanities-txt.}.
Words were lemmatized, and the most and least frequent words discarded.
The final humanities collection contained 149,734 documents and the sciences collection had
160,680 documents, with
a combined vocabulary of 21,513 unique words.
Together, these collections typify a real-world situation where there is likely some,
but not overwhelming correlation.

The results indicate that the sciences and humanities share several topics. Both exhibit an interest in a ``non-human'' theme (common topic \#2; Table~\ref{tab:topwords}). This topic is quite similar in both collections (\textit{pig} and \textit{monkey} for science documents; \textit{bird} and \textit{gorilla} for humanities documents), while their shared component forms a cohesive topic (\textit{animal}, \textit{specie}, and \textit{monkey}). This kind of correlation is also evident in topic \#23, about physics. While the science documents clearly represent research in particle physics, it is interesting to find the topic is also represented by humanist research focused on cultural representations of science. This reflects a growing interest in science and technology studies that has gained recent traction in the humanities. Despite their differences, both collections engage with a similar theme, seen in the shared component with words like particle, \textit{energy} and \textit{atom}.

The results also indicate that while sciences and humanities documents can share themes, they often diverge in how they are discussed. For example, common topic \#21 could be identified as \textit{economic} or \textit{capitalist}, but in the collection-specific components, the two disciplines differ in their articulatation. Science uses terms like \textit{price} and \textit{market}, indicating an acceptance of free-market capitalism (especially as it affects the practice of science), while the humanities, which has long been critical of free-market capitalism, uses terms like \textit{rural} and \textit{community}, highlighting cultural facets of modern economics. These results provide evidence about how ideas move between the sciences and humanities --- a phenomenon that constitutes a growing area of research for historians \cite{Galison,Canales}. C-LDA provides empirical, measurable, and reproducible evidence of the shared research between these disciplines, as well as how concepts are articulated.

\section{Discussion}
\label{discussion}
Our models provide a robust way to explore large and potentially weakly-related text collections without imposing assumptions about the data. Like ccLDA and TPYP, our models account for topic-word variation at the collection level. The models accommodate asymmetry in the numbers of topics (set in C-LDA, fit in C-HDP) and provide an efficient inference method which allows them to fit data with large values for $K$, which can help find correlations in less prevalent topics.
Our primary contribution is our models' ability to accommodate asymmetries between arbitrary collections. JSTOR, the world's largest digital collection of humanities research, was an ideal application setting given the size, asymmetry, and comprehensiveness of the humanities collection. As we show, humanities and science research exhibit asymmetries with regard to vocabulary and topic structure --- asymmetries that would be systematically overlooked using existing models. By characterizing common topics as mixtures of shared and collection-specific components, we can capture a kind of topic-level homophily, where similar themes are articulated in different ways due to word-, document-, and collection-level variation. Future work on these models could explore methods to fit non-common topics for both collections. In general, C-LDA and C-HDP can be used whenever documents are sampled from ostensibly different populations, where the nature of the difference is unknown.

\section*{Acknowledgements}
Thanks to David Blei for advice on applications of the model. This work contains analysis of private, or otherwise restricted data, made available to James Evans and Eamon Duede by ITHAKA (JSTOR), the opinions of whom are not represented in this paper. Jaan Altosaar acknowledges support from the Natural Sciences and Engineering Research Council of Canada. This work was supported by a grant from the Templeton Foundation to the Metaknowledge Research Network and by grant \#1158803 from the National Science Foundation.

\bibliographystyle{acl}
\bibliography{reference}

%
%
%
%
%
%

\end{document}